\documentclass{article}

\usepackage{dsfont}
\usepackage{overpic}
\usepackage{contour}
\contourlength{1pt}
\usepackage{xspace}
\usepackage{sidecap}
\usepackage{rotating}
\usepackage[normalem]{ulem}
\usepackage{amsmath}
\usepackage{xcolor,colortbl}
\usepackage{tabularx}
\usepackage{soul}
\usepackage{url}
\usepackage{amsfonts}
\usepackage{comment}
\usepackage{booktabs}
\usepackage{etoolbox}

\newlength{\indentlaenge}
\usepackage[sglblindworkshop, final]{latinx_2025}
\usepackage{listings}
\usepackage[utf8]{inputenc} 
\usepackage[T1]{fontenc}    
\usepackage{hyperref}       
\usepackage{url}            
\usepackage{booktabs}       
\usepackage{amsfonts}       
\usepackage{nicefrac}       
\usepackage{microtype}      
\usepackage{xcolor}         

\title{De-rendering, Reasoning, and Repairing Charts with Vision-Language Models}

\author{
    \textbf{Valentín Bonás}$^{1}$ \hspace{0.1cm}
    \textbf{Martín Sinnona}$^{1,3}$ \hspace{0.1cm} 
    \textbf{Viviana Siless}$^{1}$ \hspace{0.1cm} 
    \textbf{Emmanuel Iarussi}$^{1,2}$ \hspace{0.1cm}\\
    Universidad Torcuato Di Tella, Buenos Aires, Buenos Aires, Argentina $^{1}$\\
    Consejo Nacional de Investigaciones Científicas y Técnicas, Buenos Aires, Argentina $^{2}$ \\
    Facultad de Ciencias Exactas y Naturales \\
    Universidad de Buenos Aires, Buenos Aires, Argentina $^{3}$ \\
    \texttt{vbonas@mail.utdt.edu}
}

\begin{document}

\maketitle

\begin{abstract}
Data visualizations are central to scientific communication, journalism, and everyday decision-making, yet they are frequently prone to errors that can distort interpretation or mislead audiences. 
Rule-based visualization linters can flag violations, but they miss context and do not suggest meaningful design changes.
Directly querying general-purpose LLMs about visualization quality is unreliable: lacking training to follow visualization design principles, they often produce inconsistent or incorrect feedback.
In this work, we introduce a framework that combines chart de-rendering, automated analysis, and iterative improvement to deliver actionable, interpretable feedback on visualization design. 
Our system reconstructs the structure of a chart from an image, identifies design flaws using vision-language reasoning, and proposes concrete modifications supported by established principles in visualization research.
Users can selectively apply these improvements and re-render updated figures, creating a feedback loop that promotes both higher-quality visualizations and the development of visualization literacy.
In our evaluation on 1,000 charts from the Chart2Code benchmark, the system generated 10,452 design recommendations, which clustered into 10 coherent categories (e.g., axis formatting, color accessibility, legend consistency). 
These results highlight the promise of LLM-driven recommendation systems for delivering structured, principle-based feedback on visualization design, opening the door to more intelligent and accessible authoring tools.

\end{abstract}

\section{Introduction}

We increasingly communicate through data, and charts are one of the primary vehicles for conveying it. Yet the effectiveness of charts depends not only on the data they encode but also on the design choices that shape interpretation. Poor or misleading designs can distort public opinion, bias scientific conclusions, and erode trust in data. Misinterpretations arise both from limited visualization literacy and from deliberate manipulation. A system that can automatically diagnose such issues and suggest principled improvements would help ensure that data visualizations remain accurate, clear, and trustworthy across domains—from science and policy to journalism and education.

Current tools remain limited. Rule-based visualization linters provide partial relief by flagging violations of design guidelines, but they are rigid, bounded by their encoded rules, and rarely offer actionable or well-justified alternatives. General-purpose large language models (LLMs) struggle when reasoning directly over chart images, as they lack access to explicit structural representations of the visualization. Vision–language models (VLMs) extend this capability by processing visual input, but they are typically trained on broad image–text corpora and are not grounded in visualization design literature, which limits the specificity and reliability of their feedback. These shortcomings highlight the need for an approach that combines structured design knowledge with contextual reasoning to generate principled, concrete, and user-relevant chart improvements.

Recent progress has advanced two largely separate tracks. In the LLM/VLM literature, chart de-rendering (data and structure recovery, chart-to-code) and chart question answering have made plots parseable from images. In the visualization community, rule-based linters and constraint- or ranking-based recommenders formalize best practices over programmatic specifications. Neither line, however, delivers principled critique directly from images with context-aware justifications and actionable edits.

We introduce a chart assistant framework that ingests chart images, reconstructs their underlying structure, and evaluates them. Unlike prior systems, our tool first converts the image into a structured, knowledge-rich representation, generates actionable recommendations, and translates them into re-renderable edits that users can iteratively apply in a human-in-the-loop workflow.

We evaluated our tool on 1,000 charts across common types (bars, lines, scatters) in both 2D and 3D, generating over 10,000 recommendations. Embedding and clustering these suggestions produced 10 groups with a Davies–Bouldin score of 3.30. These clusters aligned with coherent issue categories—axis formatting, color accessibility, legend consistency, and text readability, among others—showing that the system captures meaningful design principles rather than isolated critiques.

\section{Related Work}

Misuse of visualization—stemming from poor design choices, technical errors, or deliberate deception—is well documented. 
As early as the 1950s, Huff’s How to Lie with Statistics catalogued tactics such as truncated axes and distorted proportions \cite{huff2023lie}. 
Similar problems persist across journalism and scientific communication, where overloaded or ambiguous encodings can reduce comprehension and bias interpretation. Detecting and correcting such issues remains an enduring challenge.

Detecting chart content is the first step toward remediation. 
Early reverse-engineering systems established pipelines for chart detection, segmentation, and OCR—e.g., ChartSense, which coupled visual parsing with interactive extraction to recover axes, marks, and text \cite{jung2017chartsense}. 
More recent methods advance full de-rendering from images: ChartOCR extracts data tables and annotations \cite{luo2021chartocr}, ChartDETR detects heterogeneous chart elements with a unified detector \cite{xue2023chartdetr}, and ChartCoder translates chart images into executable plotting code \cite{zhao2025chartcoder}. 
Collectively, these approaches recover structure, values, and even code, but they do not evaluate design choices or recommend improvements grounded in visualization principles.

Beyond de-rendering, a complementary line of work encodes design knowledge to critique and improve charts. 
Early \emph{linting} frameworks formalized the idea of automated guidance for visualization design \cite{mcnutt2018linting}. 
Draco represents guidelines as logical constraints and uses them to score or synthesize designs that satisfy specified criteria \cite{moritz2018formalizing}. 
VizLinter extends this notion to existing Vega-Lite specifications, flagging common missteps and offering predefined fixes for each violation \cite{chen2021vizlinter}. 
While these systems make design knowledge actionable, they typically operate on programmatic specs rather than images, are bounded by the rules they encode, and provide limited contextual reasoning or justification—they rarely explain why an alternative is preferable given task, audience, or domain. 
These constraints motivate approaches that couple rule-based checks with models capable of context-aware analysis and user-guided revision.

Beyond rule-based approaches, researchers have very recently begun probing large language models and their multimodal variants for chart critique and repair. GPT-4V, for instance, can regenerate plotting code from chart images, especially under chain-of-thought prompting \cite{zhang2024gpt}. 
However, systematic evaluations paint a more cautious picture: on standardized visualization literacy tests, general-purpose LLMs/VLMs handle basic reading but struggle to identify misleading features and to ground feedback in design principles \cite{pandey2025benchmarking,hong2025llms,lo2024good}. 
Even text-only setups often yield generic, context-insensitive advice \cite{kim2023good}. 
These limitations suggest that principled critique requires more than de-rendering or prompting, mechanisms that couple literature-aware rules with vision–language reasoning and keep users in the loop to steer and validate revisions.

\section{Framework}

\begin{figure}
    \centering
    \includegraphics[width=1\linewidth]{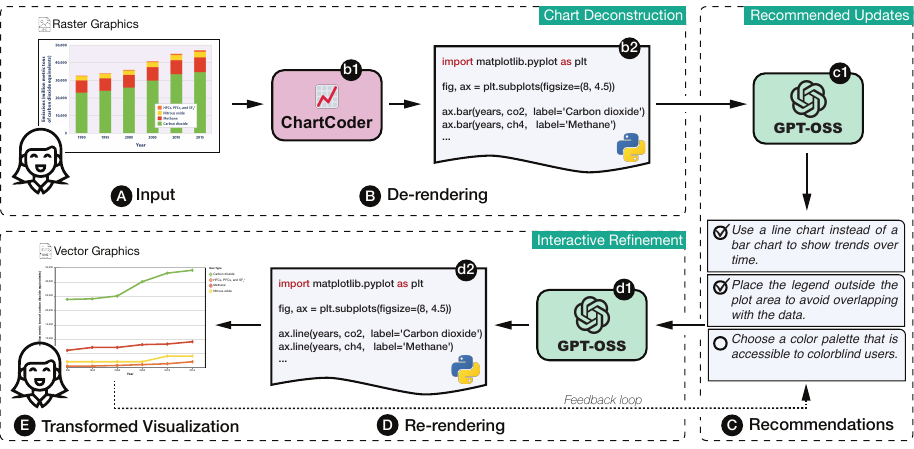}
    \caption{\textbf{System workflow overview.} Starting from a raster bar chart (A), ChartCoder de-renders the figure into executable Matplotlib code (B; b1–b2)—the \textbf{chart deconstruction} stage. GPT-OSS then inspects the recovered code and provides \textbf{recommended updates} (C; c1), e.g., switching to a line chart to show trends and placing the legend outside the plot area. Finally, through \textbf{interactive refinement} (D; d1–d2), the code is iteratively edited and re-rendered, producing a clean vector visualization with the requested changes (E).}
    \label{fig:overview}
\end{figure}

Our proposed chart assistant operates in three stages: 1- \textbf{chart deconstruction}, 2- \textbf{recommended updates}, and 3- \textbf{interactive refinement} (see Figure~\ref{fig:overview}). In \textbf{chart deconstruction}, a chart image is de-rendered to recover its structure—chart type, axes and scales, legends, marks, textual elements, and, when available, the underlying data—yielding an intermediate specification suitable for downstream reasoning. During \textbf{recommended updates}, the recovered specification is evaluated by state-of-the-art open-source LLMs, which produce ranked suggestions accompanied by concise, literature-grounded rationales. In \textbf{interactive refinement}, selected recommendations are translated into concrete edits (e.g., encodings, scales, annotations, chart type) and the figure is re-rendered, completing an interactive loop that supports iterative improvement.

\subsection{Chart Deconstruction}
The first stage of the framework recovers the structural and semantic information from the input visualization. While charts can be retained as raster images, prior work shows that structured representations enable richer reasoning and reuse than raw pixels. Code-based approaches, such as ChartCoder~\cite{zhao2025chartcoder}, translate charts into executable plotting code, offering a precise, lossless description of visual encodings that large language models (LLMs) can directly manipulate. These results suggest that structured intermediate formats substantially improve downstream tasks such as chart question answering by exposing explicit semantics rather than low-level visual features.

A key design choice is therefore the representation of extracted information. Options include detailed JSON schemas, as in PlotQA~\cite{methani2020plotqa} and FigureQA~\cite{kahou2017figureqa}, or simpler vector-based encodings. Each offers trade-offs in expressiveness, interpretability, and compatibility with LLMs.
In our system, we adopt Python plotting code as the intermediate representation. 
This decision is motivated by two considerations. 
First, most LLMs are already highly familiar with Python syntax and common visualization libraries, thereby reducing the need for models to learn a novel representation. 
Second, and more critically, we leverage ChartCoder~\cite{zhao2025chartcoder}, a dedicated Chart-to-Code multimodal large language model (MLLM). 
ChartCoder has demonstrated state-of-the-art performance on Chart2Code benchmarks, surpassing existing open-source MLLMs. 
Building upon this model allows us to generate faithful and executable chart specifications, which serve as a robust foundation for subsequent stages of automated critique and improvement.

\subsection{Recommended updates}

With the chart information represented in executable Python code, the next stage involves analyzing this specification to identify potential design issues. For this purpose, we employ \textit{Ollama}, a framework that enables flexible selection among multiple open-source language models. The selected model processes the code and outputs a structured list of visualization flaws.

Throughout development, we primarily relied on \textit{GPT-OSS 20B}, a very recent lightweight model that can be executed locally. This model consistently outperformed comparable LLaMA-based alternatives with superior prompt adherence and higher-quality recommendations. We also experimented with \textit{Gemma3’s 12B}, but at the end OpenAI's model had the better recommendations.

To guide the models, we adopt a concise, domain-specific prompt designed to elicit clear and actionable feedback. The prompt explicitly instructs the model to ignore coding or technical errors and focus solely on visual design issues. Each issue must be expressed as a single line, formatted as a comment for readability and subsequent parsing:

\begin{quote}
\raggedright\texttt{You are an expert in data visualization. Analyze the chart produced by the following Python code. Identify all visual design issues---ignore any coding or technical errors. List each issue clearly and concisely, using exactly one line per issue. Each line must begin with \#, and there should be no enumeration, explanations, or extra formatting.}
\end{quote}

This structured approach ensures that the analysis yields compact, unambiguous recommendations, which can then be ranked and refined in subsequent stages of the framework.

\subsection{Interactive refinement}

Once recommendations are generated, users can selectively decide which modifications to apply to their visualization. 
This design choice enables a flexible, human-in-the-loop workflow, ensuring that improvements align with user intent rather than being imposed automatically.

After the user confirms a subset of recommendations, the system translates these into concrete code edits and re-renders the updated visualization. 
The revised chart is then re-analyzed, producing a new round of recommendations. 
This iterative loop allows users to progressively refine their charts, balancing automated guidance with human judgment, and ultimately promotes both improved visualization quality and greater visualization literacy.

\begin{figure}
    \centering
    \includegraphics[width=1\linewidth]{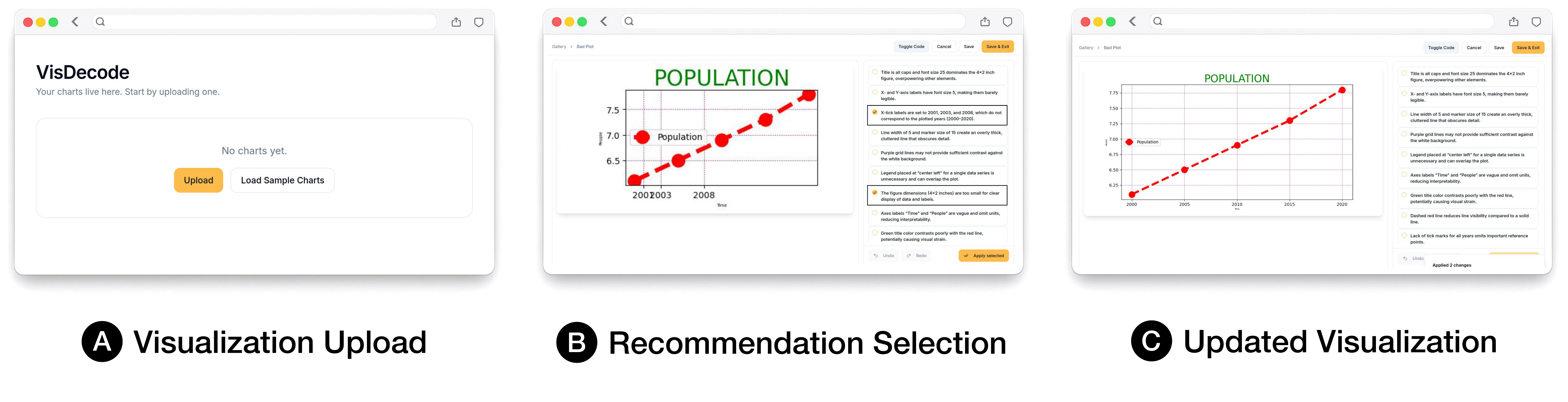}
    \caption{\textbf{Tool flow and web interface overview.} Users begin by \textbf{uploading a visualization} (A). The system then analyzes the input and presents \textbf{recommendation options} through the web UI (B), allowing users to explore alternative encodings and design choices. Finally, the selected modifications are applied, and the tool generates an \textbf{updated visualization} (C) reflecting the improvements.}

    \label{fig:placeholder}
\end{figure}

\section{Evaluation}

Our evaluation focused on assessing the quality and organization of the system’s recommendations. 
We sampled 1,000 chart images from the Chart2Code dataset, a benchmark specifically designed for evaluating chart-to-code translation. 
Each image was de-rendered into Python code using ChartCoder, yielding structured representations suitable for subsequent analysis.
From these code specifications, we generated design recommendations. 
While we experimented with multiple models, including Gemma3-12B and GPT-OSS:20B, we ultimately conducted this evaluation with GPT-OSS, which consistently produced higher-quality feedback. 
A more detailed comparative study of model performance is reported in a separate submission currently under review.

\begin{figure}
    \centering
    \includegraphics[width=1\linewidth]{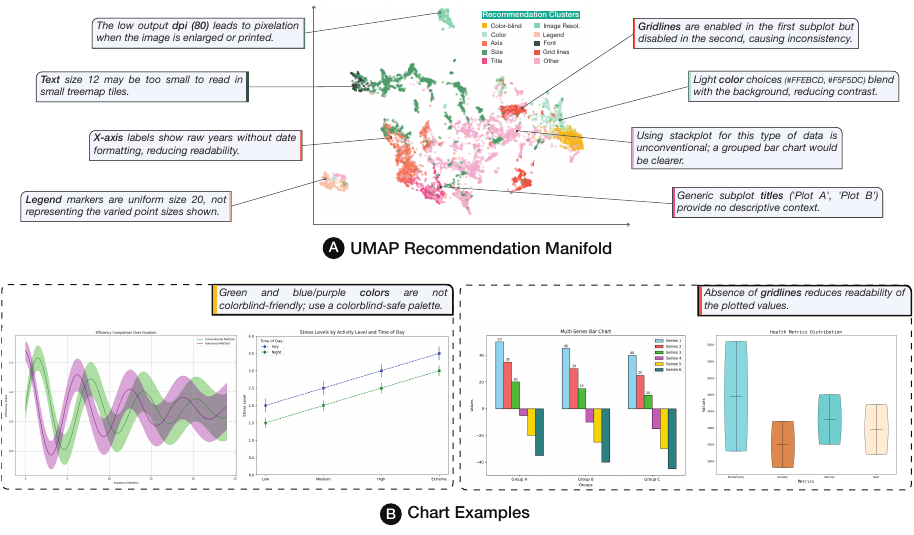}
    \caption{\textbf{Recommendation embedding space.} Visualization recommendations are encoded into feature vectors and projected with UMAP to form a \textbf{recommendation manifold} (A). Similar issues cluster together, such as low image resolution, axis formatting, or colorblind-unfriendly palettes. Example visualizations (B) illustrate specific cases, including \textbf{legend inconsistencies}, \textbf{small text sizes}, and \textbf{inadequate color contrast}, highlighting how the system groups and presents related feedback.}

    \label{fig:evaluation}
\end{figure}

To analyze the space of recommendations, each suggestion was embedded into a 1,536-dimensional vector representation using the ChatGPT API. 
This process yielded a matrix of 10,452 embeddings, which we clustered to reveal semantic groupings of related feedback.
The resulting projection (Figure~\ref{fig:evaluation}) produced ten coherent clusters, as determined by minimizing the Davies–Bouldin index. 
These clusters represent meaningful rather than random groupings. 
For instance, one cluster captured issues related to axis formatting and labeling (e.g., raw years without formatting, inconsistent gridlines), while another focused on color accessibility (e.g., non–colorblind-safe palettes, low contrast between marks and background). 
Additional clusters highlighted themes such as legend inconsistencies, font size and readability, and image resolution.

This structure indicates that the recommendations reflect underlying visualization principles, with semantically similar critiques naturally grouping together. 
In practice, this organization prevents users from being overwhelmed by a flat list of issues, instead presenting feedback as structured categories of improvements. Moreover, the clarity of these clusters suggests that the model is not merely detecting superficial flaws but is learning meaningful relationships among different aspects of visualization design.

\section{Conclusions and Future work}

We introduced the first framework that unifies automated chart deconstruction, critique, and refinement by integrating ChartCoder for chart-to-code translation with large language models for design analysis. To our knowledge, this is the only system that directly closes the loop from chart images to executable specifications, principled critique, and iterative refinement. In evaluation on 1,000 charts from the Chart2Code benchmark, the framework generated 10,452 design recommendations that naturally organized into 10 coherent categories (e.g., axis formatting, color accessibility, legend consistency). These results indicate that the system not only produces structured, principle-based feedback but also captures meaningful dimensions of visualization literacy, offering a promising direction for automated chart improvement.


Nevertheless, the system remains at an early stage and several avenues for improvement remain. 
First, while ChartCoder performs well on charts derived from code, it struggles with visualizations that originate from scanned documents, publications, or other non-programmatic sources. 
Addressing this limitation will require either developing a dedicated chart OCR pipeline or fine-tuning ChartCoder on a more diverse set of visualizations, including those from datasets such as ChartQA~\cite{masry2022chartqa}.
Second, an essential next step is to ground the system’s recommendations more explicitly in visualization literacy. 
Ensuring that critiques are interpretable and pedagogically sound would provide users with confidence that applying the suggested modifications improves not only aesthetic aspects but also the clarity and accuracy of their visualizations.
Finally, a key direction for future work is to conduct a controlled user study with visualization experts and practitioners, evaluating both the correctness and practical utility of the system’s recommendations in real-world tasks.

\textbf{Acknowledgments}. This project was supported by Universidad Torcuato Di Tella,
Argentina and Alfred P. Sloan Foundation, Grant G-2024-22665.

\bibliographystyle{plain}
\bibliography{Ref}

\end{document}